\documentclass[runningheads]{llncs}
\usepackage[T1]{fontenc}
\usepackage{algorithm}
\usepackage{amsmath}
\usepackage{amssymb} %
\usepackage{graphicx}
\usepackage{subcaption}
\usepackage{algpseudocode}
\usepackage{booktabs}
\usepackage[misc]{ifsym}

\newcommand{\corr}{(\Letter)}

\begin{document}
\title{Mixture Experts with Test-Time Self-Supervised Aggregation for Tabular Imbalanced Regression}
\titlerunning{Test-Time Aggregated Experts for Tabular Imbalanced Regression}
\author{Yung-Chien Wang \and Kuang-Da Wang \and Wei-Yao Wang \and Wen-Chih Peng \corr}
\authorrunning{Y.-C. Wang et al.} %
\institute{National Yang Ming Chiao Tung University, Hsinchu, Taiwan \\
\email{benwang.cs11@nycu.edu.tw, gdwang.cs10@nycu.edu.tw, sf1638.cs05@nctu.edu.tw, wcpeng@cs.nycu.edu.tw}}

\maketitle              %
\begin{abstract}
Tabular data serve as a fundamental and ubiquitous representation of structured information in numerous real-world applications, e.g., finance and urban planning.
In the realm of tabular imbalanced applications, data imbalance has been investigated in classification tasks with insufficient instances in certain labels, causing the model's ineffective generalizability.
However, the imbalance issue of tabular regression tasks is underexplored, and yet is critical due to unclear boundaries for continuous labels and simplifying assumptions in existing imbalance regression work, which often rely on known and balanced test distributions. Such assumptions may not hold in practice and can lead to performance degradation. 
To address these issues, we propose MATI: Mixture Experts with Test-Time Self-Supervised Aggregation for Tabular Imbalance Regression, featuring two key innovations: (i) the Region-Aware Mixture Expert, which adopts a Gaussian Mixture Model to capture the underlying related regions.
The statistical information of each Gaussian component is then used to synthesize and train region-specific experts to capture the unique characteristics of their respective regions.
(ii) Test-Time Self-Supervised Expert Aggregation, which dynamically adjusts region expert weights based on test data features to reinforce expert adaptation across varying test distributions. 
We evaluated MATI on four real-world tabular imbalance regression datasets, including house pricing, bike sharing, and age prediction, covering a range of narrow to wide target distributions. 
To reflect realistic deployment scenarios, we adopted three types of test distributions: a balanced distribution with uniform target frequencies, a normal distribution that follows the training data, and an inverse distribution that emphasizes rare target regions. 
On average across these three test distributions, MATI achieved a 7.1\% improvement in MAE compared to existing imbalanced regression methods, along with extensive analyses of the effects of each module.
\end{abstract}

\section{Introduction}
The exploration of tabular applications has broad applicability across numerous domains.
Whether it involves property valuations \cite{DBLP:conf/cikm/DuWP23,DBLP:conf/pakdd/LiWDP24}, urban planning \cite{yigitcanlar2020artificial,wang2023automated}, or fraud detection \cite{alfaiz2022enhanced,lebichot2021incremental}, these scenarios can be effectively framed as tabular classification/regression tasks characterized by heterogeneous and non-structural relations between samples (rows) and features (columns).
Data imbalance has been a significant hurdle in real-world tabular applications, leading to biased predictions and unreliable outcomes.
For instance, imbalances occur in human age estimation, where rare instances (e.g., newborn babies) tend to be influenced by the model's prior knowledge of the majority class (e.g., young adults), resulting in a negative impact on performance \cite{yang2021delving}. 
In recent years, the advancements in mitigating data imbalance for tabular data have been investigated for classifications \cite{yang2021delving,ren2022balanced,zhang2022self} as well as regression tasks \cite{yang2021delving,steininger2021density,ren2022balanced}.  

\begin{figure*}
  \begin{subfigure}[b]{0.42\textwidth}
    \includegraphics[width=\textwidth]{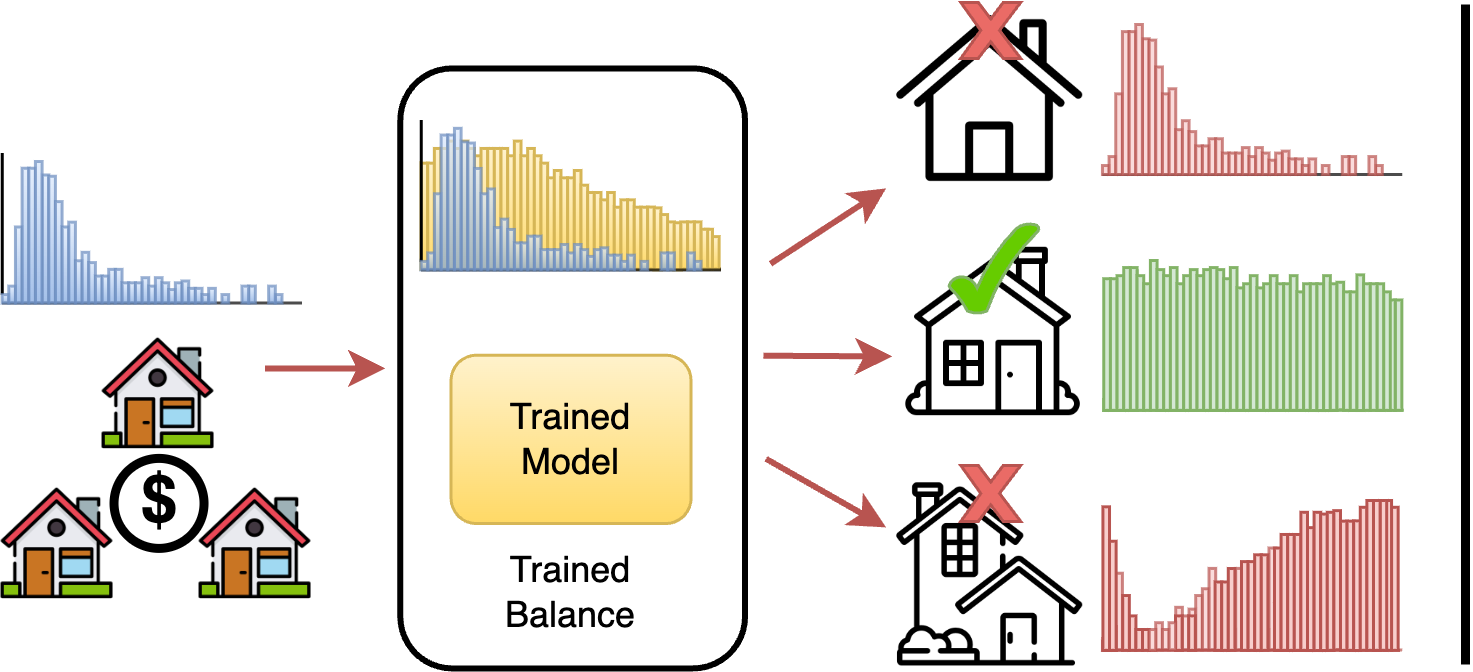}
    \caption{Existing imbalanced regression methods.}
    \label{fig:1a}
  \end{subfigure}
  \hfill %
  \begin{subfigure}[b]{0.57\textwidth}
    \includegraphics[width=\textwidth]{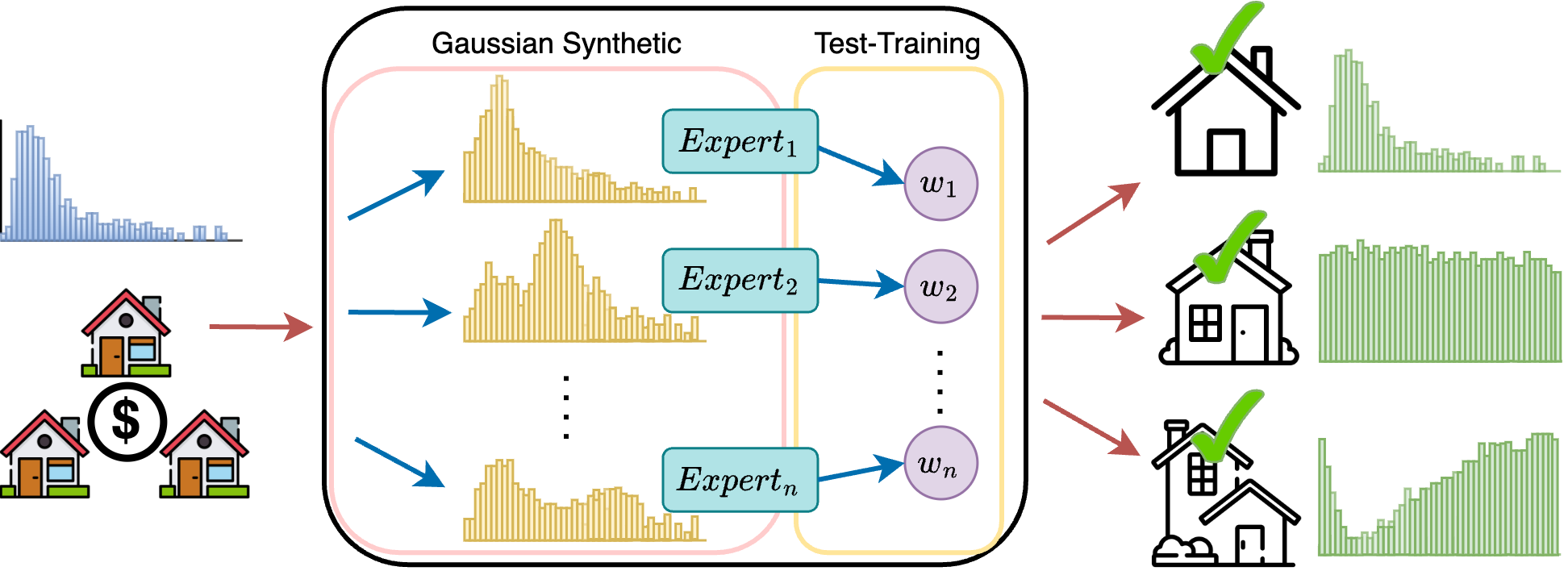}
    \caption{Our proposed method.}
    \label{fig:1b}
  \end{subfigure}
  
  \caption{Illustration of MATI. (a) Existing works on imbalanced regression focus on single test distribution which is balanced; however, there might be a performance drop when dealing with a distribution shift. (b) Our method trains \textit{n} expert models based on statistical information of Gaussian components in training data, then adjusts the expert outputs using test data features. In this manner, our method can handle different shapes of test distributions.}
  \label{fig:1}
\end{figure*}

However, these methods fall short in two critical challenges when applied to tabular imbalanced regression tasks:
First, regression labels are continuous and often unbounded, which differentiates them from classification labels that are discretized explicitly.
This challenge arises from the difficulty in learning the diverse characteristics of certain target regions, especially those that rarely appear.
Second, existing imbalanced regression methods often assume that test distributions are known and balanced \cite{yang2021delving,ren2022balanced}.
In practice, test distributions can vary significantly, and this mismatch between training and test data distributions can result in substantial performance degradation on few-shot regions.
A potential approach is to adopt \cite{zhang2022self} addressing unknown test distributions in imbalanced classification; nonetheless, these classification-focused methods cause ambiguity when transferred to regression tasks due to the dependencies between target indices \cite{yang2021delving}.
This highlights the urgent need to develop approaches for the distinct challenges of imbalanced regression datasets.

In this paper, we propose Mixture Experts With Self-Supervised Aggregation for Deep Imbalance Regression, MATI, a new approach with two novel techniques:
\textbf{1) Learning experts from imbalanced data}: We introduce the Region-Aware Mixture Expert by first dividing training data into multiple Gaussian components to acquire statistical information on each component generated by employing a regression synthesizer, and then training expert models based on synthesized datasets to focus on characteristics of specific regions. 
\textbf{2) Self-adjusting expert weights at test-time}: We introduce Test-Time Self-Supervised Expert Aggregation by perturbing test inputs \cite{yoon2020vime} and adjusting expert weights based on unlabeled test data.
By minimizing the prediction gap of expert models on perturbed and unperturbed data, expert models with more expertise receive larger weights. 
To verify the relation between continuous prediction gap minimization and regression tasks, we extend the theoretical correlation between expert contribution adjustment and prediction stability from classification to regression.

To comprehensively study imbalance performance on each target distribution, we extend the DIR benchmark \cite{yang2021delving} from balanced distributions to include two additional imbalanced test sets, a \textit{normal} test set and an \textit{inverse} test set, as shown in Figure \ref{fig:1}.
These test sets represent different imbalance levels in regression tasks, with the normal distribution following the training distribution and the inverse distribution following the reciprocal of the normal.
We then extensively evaluated our MATI and state-of-the-art baselines on four benchmarks in normal, balanced, and inverse distributions.
Our results demonstrate a notable improvement of 7.1\% in terms of average MAE scores and illustrate that existing methods significantly degrade their performance on normal and inverse distributions, indicating the critical importance of integrating diverse distributions into the testing data for evaluations.
We highlight our main contributions as follows:
\begin{itemize}
\item We propose a novel tabular regression model via Mixture Experts with Self-Supervised Aggregation, MATI, for tabular imbalance regression. To the best of our knowledge, this is the first work that tackles varying imbalances across test distributions in regression tasks.
\item MATI consists of region-aware mixture experts to enable expert models to focus on corresponding distributions based on synthetic datasets, and test time self-supervised expert aggregation to dynamically adjust expert weights based on the distributions of test data.
\item We introduce a novel evaluation approach for tabular imbalanced regression tasks incorporating additional normal and inverse test sets, providing nuanced assessments across diverse imbalanced distributions. Experimental results show that MATI outperforms existing methods with an average MAE improvement of 7.1\% across these test distributions.
\end{itemize}

\section{Related Tabular Imbalanced Works}
\label{gen_inst}

\begin{figure*}
  \includegraphics[width=\linewidth]{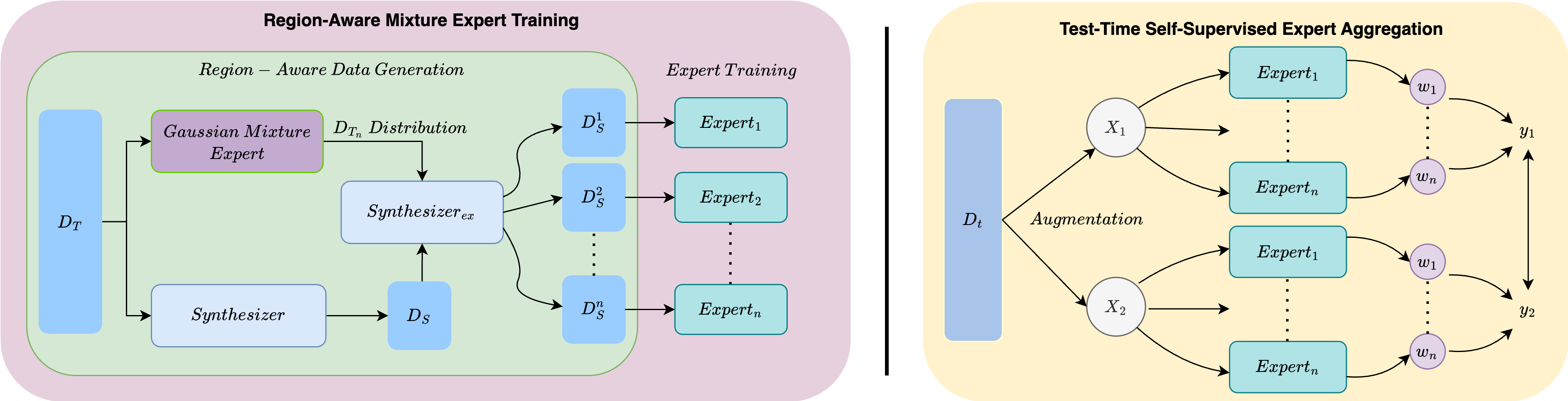}
  
  \caption{Method overview of MATI, where $D_T$ denotes the imbalanced training data, $D_S$ denotes the dataset after overall synthesizing. $D_S^n$ denotes the datasets after the second synthesizing that considered statistical information from each Gaussian component. The expert training part in purple trains region-aware expert models. The expert aggregation part in yellow aggregates the expert models with unlabeled test data $D_t$. }
  \label{fig:2}
\end{figure*}

\subsection{Imbalanced Classification}
Existing techniques for handling imbalanced data mainly focus on imbalanced and long-tailed classification, including rebalancing, logit adjustment, feature enhancement, and ensemble methods. Rebalancing comprises resampling \cite{torgo2013smote,chu2020feature,he2009learning,kim2020m2m,guo2021long,park2022majority} and reweighting \cite{cao2019learning,cui2019class,huang2016learning,jamal2020rethinking,kang2019decoupling}, both designed to balance classes by the corresponding techniques during the training process.
Logit adjustment methods \cite{menon2020long,tian2020posterior,ren2020balanced,peng2021optimal,li2022long} adjust the output logits based on the frequencies of training labels.
However, both rebalancing and logit adjustment methods focus on tail classes yet overlook overall class distributions, leading to increased sensitivity to tail classes and consequently, higher model variance \cite{wang2020long}.
Feature enhancement methods \cite{li2024feature,gao2024enhancing} aim to improve the representation of minority classes by integrating features across class distributions, thereby enhancing the model's ability to recognize underrepresented classes. These methods are tailored for visual data, and thus may not effectively apply to the structured nature of tabular data.

On the other hand, ensemble methods \cite{zhang2022self,Cai_2021_ICCV,wang2020long} are designed to capture diverse information (e.g., facial characteristics of ages) from imbalanced datasets and effectively aggregate this information through multi-expert approaches. These methods leverage multiple specialized models to address various aspects of the data.
Regarding ensemble approaches that are designed to be test-agnostic, LADE \cite{hong2021disentangling} utilizes test data distribution as available information to post-adjust model outputs, while SADE \cite{zhang2022self} introduces a test-time training method for imbalanced classification that adjusts expert weights based on the assumption of discrete target space. However, LADE and SADE were both designed for categorical indices, which cannot be directly adapted to imbalanced regression. 
On the flip side, our method adopts a region-aware training technique with a regression synthesizer and a self-supervised method to dynamically aggregate models for imbalanced regression. We enhance the representation of few-shot regions through multiple experts and adjust expert contributions through test-time training based on varying test distributions.

\subsection{Imbalanced Regression}
Imbalanced regression is less explored compared to classification due to the unclear boundaries of regression targets, which leads to fewer benchmark datasets for imbalanced regression.
Existing works on imbalance regression resample data by interpolating features and labels \cite{torgo2013smote} or adding Gaussian noise \cite{branco2017smogn}.
These methods of imbalanced regression are directly or indirectly related to SMOTE \cite{chawla2002smote} which was developed for classification tasks that generate synthetic samples for the minority class to balance the dataset. 
These approaches suffer from a lack of context awareness, and the fact that resampling methods highly rely on statistical properties for synthesizing \cite{ribeiro2020imbalanced}.

\cite{yang2021delving} curated a deep imbalanced regression (DIR) benchmark and proposed label smoothing using kernel density estimation to smooth and estimate label frequencies for reweighting techniques. \cite{wang2024variational} improved imbalanced regression by introducing probabilistic smoothing combined with variational inference, which enhances accuracy across underrepresented regions in the image data distribution.
For logit adjustment, \cite{ren2022balanced} adjusted the mean squared error (MSE) loss function by incorporating a balancing term, which modifies the standard MSE loss to give more weight to errors from underrepresented (minority) classes or regions in the data distribution.
However, these imbalanced regression methods assume balanced test distributions, making them inadequate for addressing test-agnostic evaluation (e.g., imbalanced distributions at test time).

\section{Methodology}
\label{headings}
In this section, we introduce the problem setting, evaluation protocols, and details of MATI's two novel designs as illustrated in Figure \ref{fig:2}: Region-Aware Mixture Expert Training and Test-Time Self-Supervised Expert Aggregation.

\subsection{Problem Formulation}
We focus on tabular imbalance regression tasks: consider a dataset $\mathcal{D} = \{(x_i, y_i)\}_{i=1}^{N}$, where each $x_i \in \mathbb{R}^d$ denotes the input features, and $y_i \in \mathbb{R}$ represents the associated continuous target. To address the challenge of imbalanced regression, we follow the definition of \cite{yang2021delving} to define the label space $Y$ as a partitioned set of $B$ non-overlapping intervals or bins, such that $Y = \bigcup_{b=1}^{B} [y_{b-1}, y_b)$.
Note that $B$ is a finite set of indices $\{1, \ldots, B\}$ corresponding to these bins, and $\delta_y = y_{b+1} - y_b$ defines the resolution of interest within the target variable.

\subsubsection{Evaluation Protocol}
Previous methods use either a balanced evaluation metric or a balanced test set to assess performance across samples with varying rarities \cite{ren2022balanced}. 
However, we argue that assuming the test distribution to be balanced is inappropriate in practice; actual test distribution in regression might be shifted and possess different levels of imbalance \cite{zhang2022self}.
To address this issue, we propose a new evaluation protocol for imbalanced performance by sampling test sets with different weights for each target bin, resulting in three kinds of test sets: balanced, normal, and inverse. 
These test sets are drawn from an imbalanced training distribution $p_{\text{train}}(x,y)$, which is skewed towards certain regions of $p_{\text{train}}(y)$. 
The balanced test set $p_{\text{bal}}(x,y)$ is uniformly distributed \cite{yang2021delving}, the normal test set $p_{\text{norm}}(x,y)$ follows the target bin frequency of $p_{\text{train}}(y)$, and the inverse test set $p_{\text{inv}}(x,y)$ follows the reciprocal distribution of the normal test set. 
Our goal is to learn a model from training data $D_T$ with a distribution $p_{\text{train}}(x,y)$ to predict effective results under balanced, normal, and inverse distribution scenarios.

\subsection{Region-Aware Mixture Expert Training}
Inspired by the successful advancements of utilizing the multi-expert techniques in vision \cite{hong2021disentangling,Cai_2021_ICCV,zhang2022self,chen2022multi} and language \cite{guo2024largelanguagemodelbased,jiang2024mixtralexperts} applications, where each expert model is specifically skilled in handling certain classes or target regions, we introduced region-aware mixture experts by extending expert models dedicated to specific distributions of imbalanced regression tasks. 
Specifically, we train our region experts by synthesizing different regions of data based on the distribution information from \( D_T \).

The process of training target region experts is depicted in the left part of Figure \ref{fig:2}.
To mitigate the imbalance by generating synthetic samples in underrepresented regions, the training set \( D_T \) undergoes a synthesizer designed based on SMOGN \cite{branco2017smogn}, which includes a minority oversampling technique tailored for tabular regression.
The synthesizer automatically defines rare regions for over-sampling within the entire target space $Y$ of \( D_T \).
This results in a dataset, denoted as \( D_S \), which is more balanced compared to the original data \( D_T \):
\begin{equation}
D_S = \text{Synthesizer}(D_T, Y).
\end{equation}

 The original data with imbalance distribution may consist of multiple complex distributions where a single model might be biased towards the nearest majority regions \cite{dong2018imbalanced}. To capture this distributional information and enable multiple models to correspond accordingly, a Gaussian Mixture Model (GMM) is applied to $D_T$ to capture the distinct distributions within the original training data.
 We divide the dataset \( D_T \) with its labels \( Y \) into multiple datasets by fitting GMM on the original dataset and its label.
 Formally, the GMM is defined as:
\begin{equation}
\text{GMM}(D_T) = \sum_{n=1}^{N} \pi_n \mathcal{N}(Y \mid \mu_n, \sigma_n),
\end{equation}
where \( N \) is the number of Gaussian components, \( \pi_n \) are the mixture weights, and \( \mathcal{N}(Y \mid \mu_n, \Sigma_n) \) is the Gaussian distribution with mean \( \mu_n \) and variance \( \sigma_n \).

Afterwards, each data point \( (x_i,y_i) \in D_T \) is assigned to a cluster \( n \) based on the posterior probability:
\begin{equation}
z_i = \arg \max_{n} \; \pi_n \mathcal{N}(y_i \mid \mu_n, \sigma_n),
\end{equation}

Generally, the original dataset \( D_T \) can be divided into \( N \) subsets \( D_{T_n} \), where each subset contains data points assigned to the \( n \)-th Gaussian component:
\begin{equation}
D_T = \bigcup_{n=1}^{N} D_{T_n}, \quad \text{where } D_{T_n} = \{ (x_i,y_i) \mid z_i\}.
\end{equation}

GMM provides a statistical approach to model the complex underlying structure of $D_T$, while it fits $D_T$ by its label $Y$ into $n$ sets of data, each denoted as $D_{T_n}$, where $n$ is decided by the Akaike Information Criterion score \cite{kuha2004aic} for better training and generalization results.
With the Gaussian components, we utilize a second synthesizing on $D_S$ based on the statistical information of each $D_{T_n}$, denoted as $\text{Synthesizer}_{ex}$ as in Figure \ref{fig:2}.
$\text{Synthesizer}_{ex}$ synthesizes data in the same manner as $\text{Synthesizer}$, except that the over-sampled regions are replaced with statistical information.
Specifically, we over-sample on a specified target range of $D_S$, and the range is decided by the label mean $ \mu_n$ and standard deviation $\sigma_n$ of each component $D_{T_n}$ with an adjustable hyperparameter $\alpha$ multiplied on $\sigma_n$:  
\begin{equation}
D_S^n = \text{Synthesizer}_{ex}(D_S, \mu_n-\alpha \sigma_S^n,  \mu_n+ \alpha \sigma_S^n).
\end{equation}
 The outcome of this training step is a collection of datasets after over-sampling a specified target range of $D_S$, denoted as $D_S^n$. The final phase involves the development of multiple expert models, represented as $Expert_n$.
 Each expert is trained on its corresponding training data $D_S^n$ with the MSE loss, ensuring that every model develops a specialized understanding of a particular target region.

\subsection{Test-Time Self-Supervised Expert Aggregation for Regression}
\begin{algorithm}[t]
    \caption{Test-Time Self-Supervised Expert Aggregation}
    \label{alg:test_time_aggregation}
    \begin{algorithmic}[1]
    \Require Epochs $T'$; The trained experts $E_1$ to $E_n$; Corrupt ratio $r$
    \State \textbf{Initialize:} Expert aggregation weights $w$ with uniform initialization
    \For{$e=1$ to $T'$}
        \For{each $x \in \text{Test}$} \Comment{batch sampling in practice}
            \State Obtain perturbation functions $t \sim \mathcal{T}$, $t' \sim \mathcal{T}$
            \State Set the perturbation ratio $r$
            \State Generate data views $x^1 = t(x, r)$, $x^2 = t'(x, r)$
            \State Obtain logits $v_1^1$ to $v_n^1$ for the view $x^1$
            \State Obtain logits $v_1^2$ to $v_n^2$ for the view $x^2$
            \State Normalize expert weights $w$ via softmax function
            \State Conduct predictions $\hat{y}^1$, $\hat{y}^2$ based on $\hat{y} = wv$
            \State Compute prediction gap $S$ \Comment{Eq. (6)}
            \State Minimize $S$ to update $w$
        \EndFor
        \If{$w_i \leq 0.05$ for any $w_i \in w$}
            \State Stop training
        \EndIf
    \EndFor
    \State \textbf{Output:} Expert aggregation weights $w$
    \end{algorithmic}
\end{algorithm}

After training region-specific experts, our next objective is to aggregate these expert models without prior knowledge of the test distribution and labels, allowing for effective adaptation to test distributions with varying degrees of imbalance.
To that end, our intuitions are that 1) there exists a positive correlation between expertise and prediction stability, i.e., when predicting different views of samples, experts have higher prediction similarity of samples from their favorable regions, and 2) the principle that a model with more expertise in a particular area should assume a more significant role and consequently carry greater weight after the aggregation.

As shown in algorithm \ref{alg:test_time_aggregation}, we aggregate the trained region experts by minimizing the continuous output gap of different views of a test sample since experts proficient in specific regions should demonstrate enhanced stability in their predictions.
Hence, \textit{Continuous Prediction Gap Minimization} is proposed to ensure that the contribution of experts is effectively adapted to test distributions with varying degrees of imbalance.

\subsubsection{Continuous Prediction Gap Minimization}
To tackle varying test distributions, we dynamically adjust region expert contributions by learning aggregation weights for frozen parameter experts by minimizing the model output gap between two similar unlabeled test samples, which is generated by adding slight perturbation on the other one. 
Let \( D_t \) denote a mini-batch of test samples used for test-time adaptation.
As shown in the right part of Figure \ref{fig:2}, the method involves three steps: 
\begin{enumerate}
    \item \textbf{Perturbation.}  For each test sample \( X \in D_t \), we generate two perturbed views \( X_1 \) and \( X_2 \) using VIME \cite{yoon2020vime}, a simple yet effective tabular augmenting approach.
    \item \textbf{Expert Inference.}The parameters of all region experts \( \text{Expert}_1, \dots, \text{Expert}_n \) are frozen. Each expert produces a logit output \( v_i \), and we assign a learnable weight \( w_i \in \mathbb{R} \) to each expert. Only the weights \( w_i \) are updated during test-time training.
    \item \textbf{Prediction Aggregation.} The weights are normalized using a softmax function \( \sigma \), and the final prediction is computed as a weighted sum of expert outputs: $y = \sum_{i=1}^{n} \sigma(w_i) \cdot v_i.$
\end{enumerate}
Applying this to both views \( X_1 \) and \( X_2 \), we obtain corresponding outputs \( y_1 \) and \( y_2 \). The goal is to minimize the prediction difference between these two views, thereby encouraging stable expert combinations for each test input.

The objective function of minimizing the prediction gap in regression is defined as:
\begin{equation}\label{eq6}
\min_w S, \quad \text{where } S = \frac{1}{|D_t|} \sum_{X \in D_t} (y_1 - y_2)^2.
\end{equation}

Optimizing the prediction gap $S$ assigns greater weights to more proficient experts, adapting to the unknown test distribution. 
This self-supervised aggregation approach facilitates expert weight adjustment without relying on test labels.

\subsubsection{Theoretical Analysis}
We theoretically adapt the prediction stability maximization strategy to a regression context. 
Given random variables predictions and labels defined as \( \hat{Y} \sim p(\hat{y}) \) and \( Y \sim p_t(y) \), our goal is to adjust the weights of experts to produce predictions \( p(\hat{y}) \) that better align with \( p_t(y) \). This can be achieved through the maximization of mutual information between the predicted distribution and the true label distribution \cite{wen2016discriminative}.
For regression, prediction gap $S$ is proportional to the mutual information between the predicted distribution \( p(\hat{y}) \) and the test distribution \(  p_t(y) \). 
This relationship is formally stated as follows:

\noindent\textbf{Theorem 1.} The prediction gap \( S \) is positively proportional to the mutual information between the predicted distribution and the test distribution \( I(\hat{Y}; Y) \):
\[
S \propto I(\hat{Y}; Y).
\]

To prove this theorem, we introduce the concept of center loss, which minimizes the distance between features and their respective class centers. 
Center loss has a demonstrated proportional relationship with mutual information \( I(\hat{Y}; Y) \) in classification tasks \cite{zhang2022self,wen2016discriminative}.
We link prediction gap \( S \) with mutual information by utilizing center loss in the regression context \cite{zhang2022self,zhang2023improving}. 
By categorizing the continuous target space into bins \cite{yang2021delving,zhang2023improving}, we can relate the MSE of prediction gap \( S \) (cf. Eq. \ref{eq6}) and center loss of these bins in regression tasks:

\begin{multline}
\sum_{\substack{y_j, y_j' \in Z_k}} -(y_j-y_j')^2 = \frac{1}{|Z_k|} \sum_{\substack{y_j, y_j' \in Z_k}} -(y_j-y_j')^2 \\
=-\frac{1}{|Z_k|} \sum_{\substack{y_j, y_j' \in Z_k}} (y_j^2 - 2y_j y_j' + y_j'^2) \\
= -\left( \sum_{y_j \in Z_k} y_j^2 - 2 \frac{1}{|Z_k|} \sum_{y_j \in Z_k} \sum_{y_j' \in Z_k} y_j y_j' + \frac{1}{|Z_k|} \sum_{y_j \in Z_k} \sum_{y_j' \in Z_k} y_j y_j' \right) \\
= -\left( \sum_{y_j \in Z_k} y_j^2 - 2 \sum_{y_j \in Z_k} y_j c_k + c_k^2 \right) = \sum_{y_j \in Z_k} \|y_j - c_k\|^2,
\end{multline}
where $|Z_k|$ denotes the number of samples in bin \( k \), \( c_k \) represents the hard mean of all predictions of samples from bin \( k \), where \( c_k = \frac{1}{|Z_k|} \sum_{\hat{y} \in Z_k} \hat{y} \), for the equations to indicate equality up to a multiplicative and/or additive constant. By introducing the strategy of \textit{Continuous Prediction Gap Minimization} to the regression scenario, we show that minimization of $S$ is proportional to the maximization of mutual information between prediction and the true label; this adjusts the weights of experts to the output prediction \( p(\hat{y}) \) that better fits \(  p_t(y) \).

\section{Experiments}
\label{others}

\subsection{Experimental Settings}
\begin{figure*}
    \centering
    \includegraphics[width=0.95\textwidth]{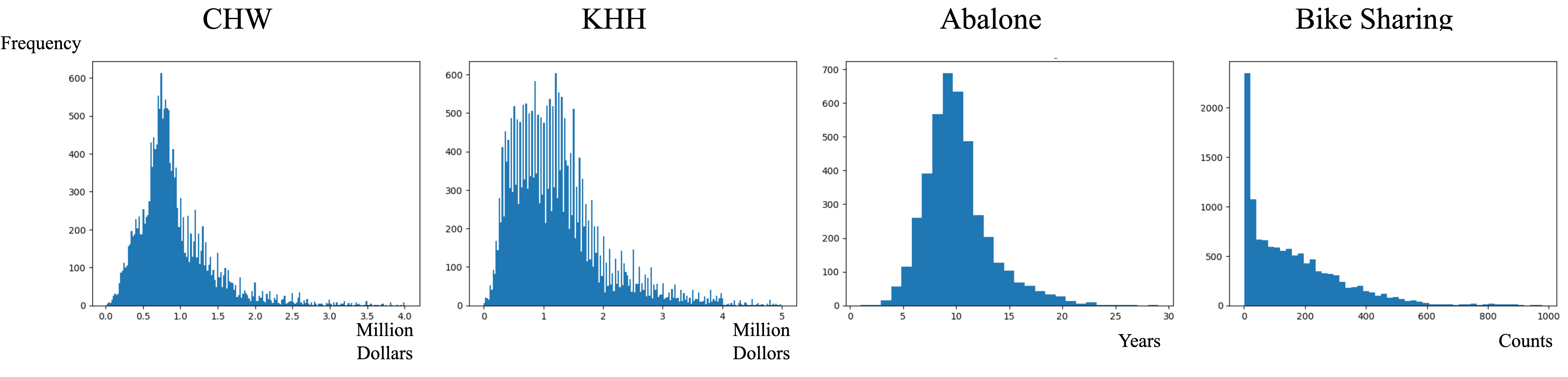}
    \caption{Distribution of the four DIR datasets.}
    \label{fig:4}
\end{figure*}
\noindent\textbf{Datasets.} 
We curated four real-world tabular imbalance regression benchmarks across diverse domains, as shown in \ref{fig:4}, including pricing, aging, and counts, and constructed three distinct test data distributions: balanced, normal, and inverse. \textbf{1) Taiwan House} \cite{DBLP:conf/pakdd/LiWDP24} focuses on house price appraisal for two Taiwanese regions, CHW and KHH, both exhibiting long-tail distributions with extremely high-price bins. CHW represents a smaller city with lower prices, while KHH corresponds to a larger city with higher average prices. We set a bin range of 1 million and appended few-shot bins with extreme values to the last bin. The task involves predicting house prices using metadata, with three components determined via AIC. \textbf{2) Abalone} \cite{misc_abalone_1} predicts abalone age from physical measurements, characterized by an imbalanced target range lacking both low and high ages. The bin size was set to 1, and two components were identified. \textbf{3) Bike Sharing} \cite{misc_bike_sharing_275} estimates bike rental counts based on weather, date, and weekday data, with a bin range of 5 to account for few-shot bins with extreme values. Three components were used. Evaluation metrics included MAE and MAPE for the CHW and KHH datasets, leveraging MAPE's suitability for large-range targets, and RMSE and MAE for Abalone and Bike Sharing to assess regression performance. These datasets provide a robust foundation for testing under diverse imbalanced conditions.

\noindent\textbf{Baselines and Implementation Details.} 
To validate the effectiveness of MATI, we benchmarked it against several state-of-the-art baselines: 1) \textbf{Vanilla}, a simple approach without imbalanced regression techniques, trains TabNetRegressor using the same settings as our expert models. 2) \textbf{SMOGN} \cite{branco2017smogn}, which applies MATI’s synthetic process across the entire target space, identifies minority regions and synthesizes data with Gaussian noise before training TabNetRegressor. 3) \textbf{RRT} \cite{yang2021delving}, which retrains only the final layer using inverse reweighting after initial full-model training. 4) \textbf{RRT+LDS}, an extension of RRT incorporating Label Distribution Smoothing (LDS) with kernel density estimation for refined weighting. 5) \textbf{BMC} \cite{ren2022balanced}, which replaces standard MSE loss with Balanced MSE Correction, leveraging label distribution priors for balanced predictions. These baselines provide rigorous evaluation for MATI’s performance.

For the baseline methods and MATI, we used TabNetRegressor \cite{arik2021tabnet} as our base model for fair comparisons. 
We trained each method for a maximum of 200 epochs, with an early stopping set of 20 epochs. The model size ranged from 3 to 5 for each method, and the results are the average of three different seeds. 
For some detailed hyperparameters, we set $\alpha$ from 1 to 2 for the synthesizer.
To ensure fairness, we fix the number of epochs and the corruption ratio for test-time training in MATI. The number of epochs is set between 20 and 50, depending on the test set size, while the corruption ratio is fixed at 0.1.

\subsection{Quantitative Results on Varying Test Distributions}
Tables \ref{tab:CHW_KHH_main} and \ref{tab:Abalone_Bike_main} present the performance of MATI and the baselines on the four datasets. 
We summarize the observations as follows:
\textbf{1) Superior MATI Performance on Balanced Test Set: }
Quantitatively, MATI outperforms all baselines on the balanced test set, achieving at least 0.339 and 1.021 lower MAPE in the CHW and KHH house price datasets, respectively. It also surpasses baselines by at least 1.05\% and 8.71\% in RMSE on the Abalone and Bike Sharing datasets. 
Despite these baselines focusing on the performance of balanced test distributions, the comparisons with our MATI not only illustrate the need for evaluating various distributions but also reveal the importance of considering different distributions via the corresponding experts.
\textbf{2) The Advantage of Test-Time Aggregation:}
We can observe that imbalanced regression baselines (i.e., RRT+LDS) perform well only on balanced evaluation protocols, while MATI excels across different distributions.
Vanilla is only superior in many-shot regions which share the same distribution as the training sets, but is deleterious in few-shot regions due to the lack of imbalanced techniques. 
In contrast, MATI equipped with Test-Time Self-Supervised Expert Aggregation adjusts expert weights for adapting to test distributions. 
This allows MATI to perform slightly worse than Vanilla on the normal test set, but to achieve optimal results on both the balanced and inverse test sets. 
This highlights the effectiveness of MATI's test-time aggregation in handling to varying test distribution.

\begin{table*}[t]
    \caption{Quantitative results for the property valuation scenarios (CHW and KHH).}
    \label{tab:CHW_KHH_main}
    \resizebox{\textwidth}{!}{%
    \begin{tabular}{lcccccccccccc}
    \toprule
    & \multicolumn{6}{c}{CHW} & \multicolumn{6}{c}{KHH} \\
    \cmidrule(lr){2-7} \cmidrule(lr){8-13}
    Method & \multicolumn{3}{c}{MAPE ↓} & \multicolumn{3}{c}{MAE ↓} & \multicolumn{3}{c}{MAPE ↓} & \multicolumn{3}{c}{MAE ↓} \\
    \cmidrule(lr){2-4} \cmidrule(lr){5-7} \cmidrule(lr){8-10} \cmidrule(lr){11-13}
    & Balanced & Normal & Inverse & Balanced & Normal & Inverse & Balanced & Normal & Inverse & Balanced & Normal & Inverse \\
    \midrule
    Vanilla & 26.221 & \textbf{21.714} & 30.362 & 0.296 & 0.334 & 0.339 & 27.556 & \textbf{21.111} & 32.333 & 0.301 & 0.343 & 0.378 \\
    SMOGN & 28.252 & 24.601 & \underline{30.361} & 0.334 & 0.285 & 0.360 & 29.454 & 25.563 & \underline{32.322} & 0.356 & 0.297 & 0.366\\
    RRT & 36.244 & 25.344 & 38.762 & 0.35 & 0.311 & 0.389 & 36.236 & 25.331& 35.224 & 0.361 & 0.323 & 0.397 \\
    RRT+LDS & \underline{25.563} & 27.563 & 32.555 & 0.321 & 0.298 & 0.365 & \underline{25.354} & 28.231 & 33.456 & \underline{0.294} & 0.314 & 0.359 \\
    BMC & 32.051 & 29.231 & 35.983 & \underline{0.296} & \textbf{0.239} & \underline{0.337} & 33.989 & 28.999 & 36.932 & 0.337 & \textbf{0.232} & \underline{0.355} \\
    \midrule
    \textbf{MATI} & \textbf{25.224} & \underline{22.793} & \textbf{28.227} & \textbf{0.293} & \underline{0.254} & \textbf{0.333} & \textbf{24.333} & \underline{22.896} & \textbf{29.348} & \textbf{0.286} & \underline{0.251} & \textbf{0.342}\\
    \bottomrule
    \end{tabular}
    }
\end{table*}
\begin{table*}[t]
    \centering
    \caption{Quantitative results for the age prediction (Abalone) and rental prediction (Bike Sharing) scenarios.}
    \label{tab:Abalone_Bike_main}
    \resizebox{\textwidth}{!}{%
    \begin{tabular}{lcccccccccccc}
    \toprule
    & \multicolumn{6}{c}{Abalone} & \multicolumn{6}{c}{Bike Sharing} \\
    \cmidrule(lr){2-7} \cmidrule(lr){8-13}
    Method & \multicolumn{3}{c}{RMSE ↓} & \multicolumn{3}{c}{MAE ↓} & \multicolumn{3}{c}{RMSE ↓} & \multicolumn{3}{c}{MAE ↓} \\
    \cmidrule(lr){2-4} \cmidrule(lr){5-7} \cmidrule(lr){8-10} \cmidrule(lr){11-13}
            & Balanced & Normal & Inverse   & Balanced & Normal & Inverse   & Balanced & Normal & Inverse   & Balanced & Normal & Inverse \\
    \midrule
    Vanilla & 3.756 & \textbf{2.614} & 4.955   & 2.787 & \textbf{1.947} & 4.093   & 92.231 & \textbf{49.382} & 92.325   & 67.454 & \textbf{36.421} & 78.655 \\
    SMOGN   & 3.561 & 2.625 & 4.588   & 2.876 & 2.099 & 3.944   & 102.413 & 61.428 & 119.937   & 73.711 & 43.564 & 91.823\\
    RRT   & \underline{3.332} & 2.669 & \underline{4.234}   & \underline{2.662} &2.101 & \underline{3.467}   & 93.107 & 49.664  & 100.567   & 66.377 & 37.537 & 79.661 \\
    RRT+LDS   & 3.831 & 2.911 & 4.885   & 2.892 & 2.204 & 3.998   & 92.264 & 49.909 & 100.916   & 66.138 & \underline{37.507} & 80.477\\
    BMC   & 3.541 & 2.919 & 4.263   & 2.878 & 2.333 & 3.629   & \underline{87.783} & \underline{48.451} & \underline{86.672}   & \underline{63.689} & 37.834 & \underline{71.781} \\
    \midrule
    \textbf{MATI}  & \textbf{3.297} & \underline{2.622} & \textbf{3.855}   & \textbf{2.641} & \underline{2.079} & \textbf{3.131}   & \textbf{86.989} & 56.201 & \textbf{81.785}   & \textbf{62.140} & 37.982 & \textbf{65.310}\\
    \bottomrule
    \end{tabular}
    }
\end{table*}
\begin{table}[t]
    \centering
    \caption{Performance of experts on corresponding regions of Bike Sharing and CHW.}
    \label{tab:region_expert}
    \resizebox{\textwidth}{!}{%
    \begin{tabular}{lcccccccccccc}
    \toprule
    & \multicolumn{6}{c}{Bike Sharing} & \multicolumn{6}{c}{CHW} \\
    \cmidrule(lr){2-7} \cmidrule(lr){8-13}
    Expert & \multicolumn{3}{c}{RMSE ↓} & \multicolumn{3}{c}{MAE ↓} & \multicolumn{3}{c}{MAPE ↓} & \multicolumn{3}{c}{MAE ↓} \\
    \cmidrule(lr){2-4} \cmidrule(lr){5-7} \cmidrule(lr){8-10} \cmidrule(lr){11-13}
            & Region 1 & Region 2 & Region 3 & Region 1 & Region 2 & Region 3   & Region 1 & Region 2 & Region 3 & Region 1 & Region 2 & Region 3 \\
    \midrule
    $Expert_1$ & \textbf{18.121} & 119.814 & 166.433 & \textbf{11.776} & 91.945 & 133.471   & \textbf{29.213} & 28.9321 & 37.122 & \textbf{0.288} & 0.239 & 0.298 \\
    $Expert_2$   & 20.781 & \textbf{94.745}  & 168.383 & 16.174 & \textbf{72.602} & 138.161   & 29.789 & \textbf{27.973}  & 36.464 & 0.334 & \textbf{0.233} & 0.290\\
    $Expert_3$   & 29.870 & 105.482 & \textbf{160.133} & 20.716 & 82.391 & \textbf{128.923}   & 31.213 & 32.245 & \textbf{35.744} & 0.328 & 0.273 & \textbf{0.282} \\
    \bottomrule
    \end{tabular}
    }
\end{table}

\subsection{Ablation Studies on varying skewed test data distribution} \label{exp:4.3}
To assess the proficiency of region experts and validate the test-time aggregation mechanism of MATI, we conducted experiments on the Bike Sharing and CHW datasets, both of which exhibit highly right-skewed distributions. For these datasets, we employed three Gaussian components, sorting experts by their means to define Region 1, Region 2, and Region 3, handled by $Expert_1$, $Expert_2$, and $Expert_3$, respectively. Region-specific test sets were created by sampling based on Gaussian means and standard deviations, enabling performance evaluation of each expert in its corresponding region.Table \ref{tab:region_expert} confirms that each expert excels within   its designated region, validating the effectiveness of Gaussian Mixture Synthesizing in modeling region-specific characteristics. Furthermore, MATI’s dynamic weight adjustment during test-time aggregation effectively aligns expert contributions to the test distribution. Table \ref{tab:weights_results} illustrates that $Expert_1$, optimized for Region 1, dominates on right-skewed normal test sets; $Expert_2$, representing Region 2, is weighted higher on balanced test sets; and $Expert_3$, trained for Region 3, is prioritized on left-skewed inverse test sets. These findings demonstrate MATI’s ability to adaptively allocate expert weights, ensuring optimal performance across diverse test scenarios.

\subsection{Effects of Perturbation Ratio}

\begin{figure}[t]
  \centering
  \includegraphics[width=\linewidth]{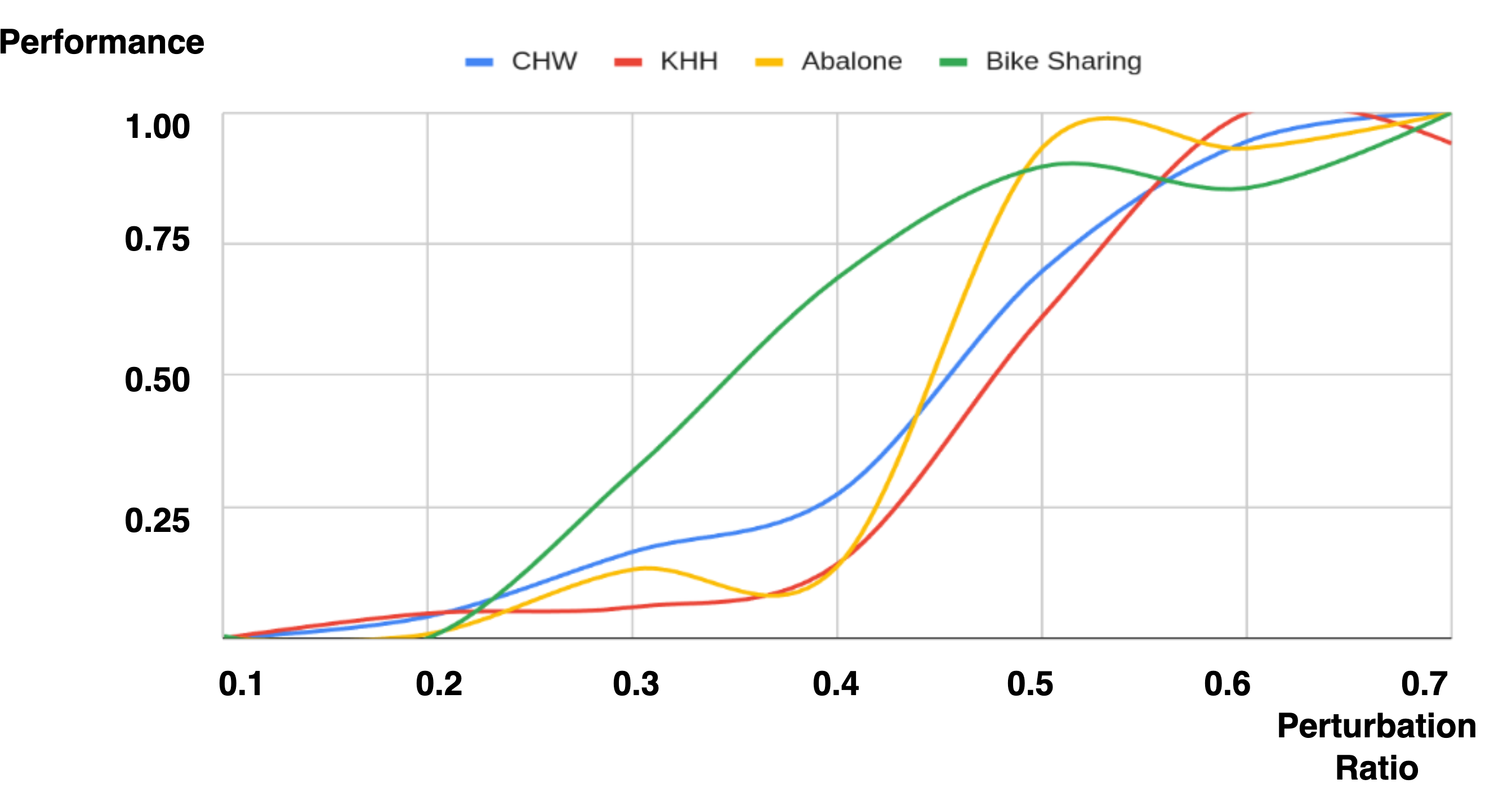}
  \caption{The relationship between the performance of the four datasets and the perturbation ratio.}
  
  \label{fig:6}
\end{figure}

To verify that the perturbation method does not generate excessively large or dissimilar samples, thereby affecting the performance, we closely examined the changes in performance with varying perturbation ratios (Algorithm \ref{alg:test_time_aggregation}, line 5) from 0.1 to 0.7.
As illustrated in Figure \ref{fig:6}, the performance curves for the four datasets remain relatively flat and stable at lower perturbation ratios. However, when the ratio exceeds 0.4, there is a noticeable and abrupt increase in the curves.
This finding indicates that a slight perturbation of the input features allows the perturbation method to generate proper and similar samples for aggregation.

\begin{table}[t]
    \centering
    \footnotesize
    \caption{Weights of experts on skewed test distributions for Bike Sharing and CHW show that MATI aggregates experts with optimal weights.}
    \label{tab:weights_results}
    \begin{tabular}{lcccccc}
    \toprule
    & \multicolumn{3}{c}{Bike Sharing} & \multicolumn{3}{c}{CHW} \\
    \cmidrule(lr){2-4} \cmidrule(lr){5-7}
    Distribution & Expert 1 & Expert 2 & Expert 3 & Expert 1 & Expert 2 & Expert 3 \\
    \midrule
   
    Right-Skewed & 0.372 & 0.334 & 0.294 & 0.400 & 0.270   & 0.330  \\
    Balanced & 0.312 & 0.443 & 0.275 & 0.310 & 0.410  & 0.280 \\
    Left-Skewed & 0.310 & 0.345 & 0.345 & 0.293 & 0.324  & 0.383 \\
    
    \bottomrule
    \end{tabular}
\end{table}

\section{Conclusion}

In this paper, we present MATI, an innovative method that combines Mixture Experts with Test-Time Self-Supervised Aggregation to address Tabular Imbalance Regression.
MATI targets the two challenges of learning from imbalanced data and adapting to varying test distributions in regression tasks.
By employing Gaussian Mixture Synthesizing for region-specific expert training and Continuous Prediction Gap Minimization for test-time aggregation, MATI enhances the representation of few-shot regions and boosts model robustness against diverse test distributions.
Distinct from existing works that only focused on balanced set evaluation, we included normal and inverse test sets for evaluations, allowing comprehensive evaluations on the capability of tackling imbalanced distributions.
Extensive evaluations of 4 real-world benchmarks show that MATI significantly outperforms state-of-the-art baselines by 7.1\% in terms of the average of various distribution and application scenarios.
We believe our MATI serves as a general framework for tabular regression applications due to the generic design for modeling imbalanced distributions as well as for the test-time aggregation technique, and multiple interesting directions could be further explored such as a unified approach for tabular imbalanced regression as well as classification applications, multi-modal features for few-shot regions, etc.

\bibliographystyle{splncs04}
\bibliography{main}

\end{document}